\pdfoutput=1

\documentclass[11pt]{article}

\usepackage[final]{acl}

\usepackage{times}
\usepackage{latexsym}
\usepackage{longtable}
\usepackage{array}
\usepackage{tabularx}
\usepackage{amsmath}
\usepackage{multirow}
\usepackage{hyperref}

\usepackage[T1]{fontenc}

\usepackage[utf8]{inputenc}

\usepackage{microtype}

\usepackage{inconsolata}

\usepackage{graphicx}

\title{DiagESC: Dialogue Synthesis for Integrating Depression Diagnosis \\ into Emotional Support Conversation}

\author{Seungyeon Seo$^1$,Gary Geunbae Lee$^{1,2}$ \\
  $^1$Graduate School of Artificial Intelligence, POSTECH, Republic of Korea\\
  $^2$Department of Computer Science and Engineering, POSTECH, Republic of Korea\\
  \texttt{\{ssy319, gblee\}@postech.ac.kr} \\
}

\begin{document}
\maketitle
\begin{abstract}
Dialogue systems for mental health care aim to provide appropriate support to individuals experiencing mental distress.
While extensive research has been conducted to deliver adequate emotional support, existing studies cannot identify individuals who require professional medical intervention and cannot offer suitable guidance.
We introduce the Diagnostic Emotional Support Conversation task for an advanced mental health management system.
We develop the DESC dataset\footnote{Our dataset DESC is accessible at github.com/seungyeon-seo/DiagESC.} to assess depression symptoms while maintaining user experience by utilizing task-specific utterance generation prompts and a strict filtering algorithm.
Evaluations by professional psychological counselors indicate that DESC has a superior ability to diagnose depression than existing data.
Additionally, conversational quality evaluation reveals that DESC maintains fluent, consistent, and coherent dialogues.

\end{abstract}

\section{Introduction}
As interest in preventing and treating mental illnesses like depression, anxiety disorders, and panic disorders grows, dialogue system studies on mental health care are gaining attention.
Several studies have shown that chatbots can effectively manage the mental health of individuals, particularly in frontline settings, before seeking professional help \citep{denecke2021artificial, LIM2022334}.
These chatbots provide emotional empathy and assist in finding stability for those facing emotional, mental, and psychological distress.
Mental health care also involves the early detection of illnesses.
Although delayed treatment aggravates symptoms and requires more complex treatment, it is challenging for individuals to self-diagnose \citep{epstein2010didn}.
Therefore, detecting diseases during conversation is an important factor, and we focus on depression, a representative mental illness.

Our research aims for an advanced conversation system to facilitate comprehensive mental health management.
This system should provide extensive emotional support to individuals while simultaneously employing diagnostic questions to detect early signs of depression proactively.
\begin{figure}[t]
  \centering
  \includegraphics[width=0.9\columnwidth]{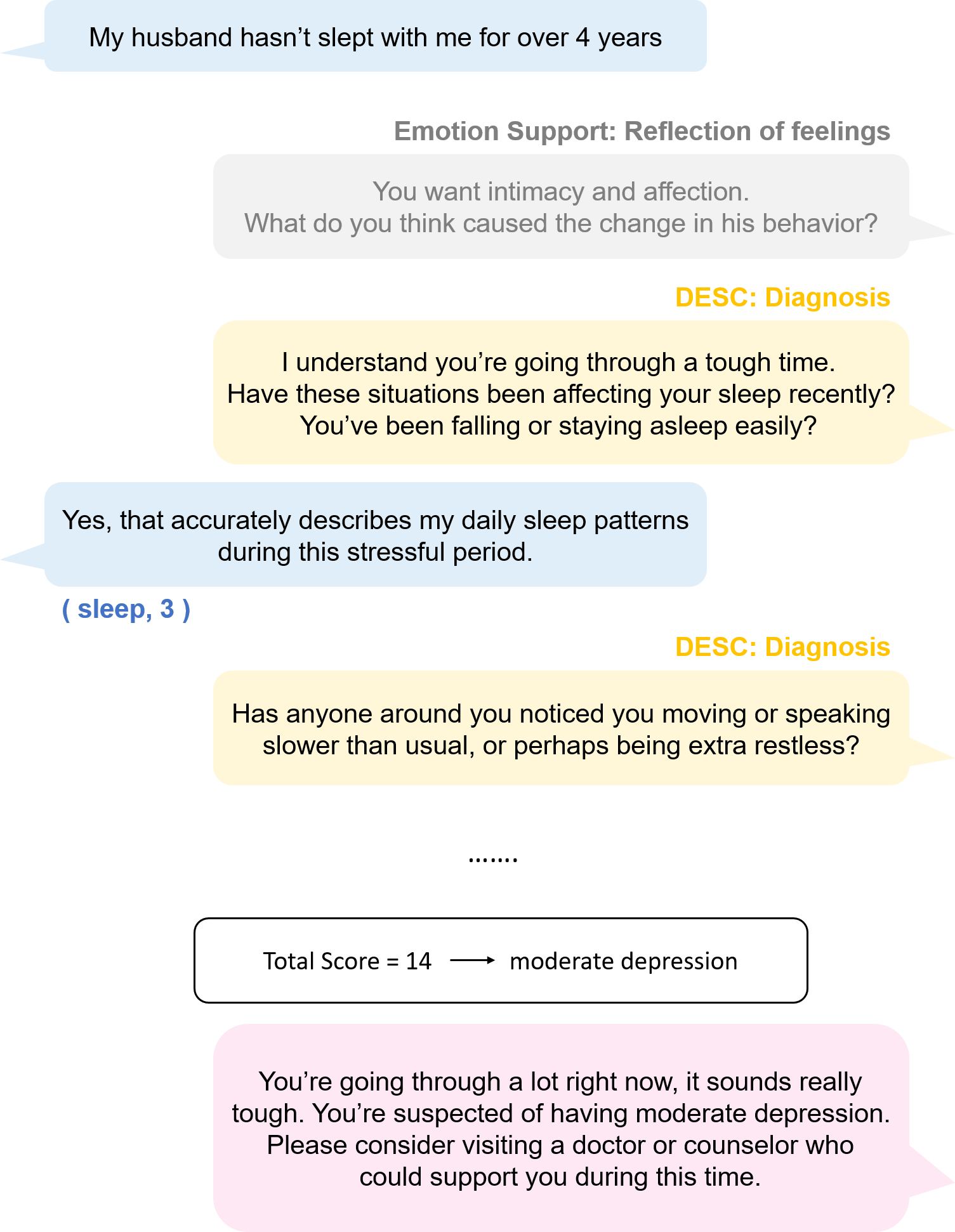}
  \caption{Part of an example conversation sample from DESC. The left is the seeker's, and the right is the supporter's utterance. We initiate a diagnostic conversation by inserting a diagnostic question (yellow) instead of a specific supporting emotion utterance (gray). At the end of the conversation, appropriate assistance (pink) is provided based on the severity of the depression.}
  \label{fig:sample}
\end{figure}

To achieve this goal, we define a novel task, Diagnostic Emotional Support Conversation (DiagESC), based on Emotional Support Conversation (ESC) \citep{liu2021towards}.
ESC aims to support by helping reduce the seeker (user)'s mental stress.
We also synthesize and release the dataset DESC for this task.
We synthesize utterances to ask questions about depression symptoms while maintaining a positive user experience.
Additionally, task-specific strict filtering algorithms ensure data quality.
Figure \ref{fig:sample} shows part of the dialogue sample in DESC.
It includes questions about depression symptoms and labels regarding symptom frequency, enabling assessing the severity of depression.
Appropriate advice based on the severity of depression helps the individual receive help.
Professional psychological counselors validate the diagnostic ability and conversational quality of DESC.

\section{Related Work}
\subsection{Supportive Dialogue System}
Recognizing emotions is essential for dialogue systems to respond appropriately to the user's feelings.
Emotion-tagged dialogue datasets such as DailyDialog \citep{li2017dailydialog}, Emotionlines \citep{chen2018emotionlines}, and EmoContext \citep{chatterjee2019semeval} have enhanced the conversation quality by enabling emotion-based response generation \citep{wei2019emotion, zandie2020emptransfo, ide2021multi}.
In particular, \citet{8649596} shows that integrating emotional context in response generation can elicit positive emotions in users.
The dataset EmpatheticDialogues \citep{rashkin2019towards} contains rich emotion labels and high-quality utterances that understand and empathize with users' emotions, encouraging research on generating empathic responses \citep{ghosal2020cosmic, majumder2020mime, Li_Li_Ren_Ren_Chen_2022}.

To enable more effective emotional support, the ESC task \citep{liu2021towards} is defined by employing response strategies based on the Helping Skills Theory \citep{Hill2009}.
ESC uses more sophisticated strategies, such as questioning and providing suggestions beyond empathy, to improve the users' emotions and encourage them to overcome difficulties.
\citet{cheng2023pal} introduced persona generation into ESC and proposed Persona-Augmented Emotional Support (PAL), enabling the creation of responses tailored to an individual's situation and characteristics.

However, understanding the situation and providing advice cannot fully help someone suffering from depression.
Individuals with depression require professional counseling and medication rather than temporary emotional support.
Research on supportive dialogue systems, such as ESC, demonstrates user encouragement capabilities but cannot adequately address the needs of those with depression.

\subsection{Depression Detection in Conversation}
As with all diseases, early detection of depression is very important for efficient treatment.
However, due to difficulties such as a lack of knowledge about the symptoms of depression, it is hard for patients to recognize that they are suffering from depression themselves \citep{epstein2010didn}.

Against this background, depression detection research is being conducted to help with early treatment.
\citet{ringeval2019avec} proposed a classification task for whether a user has depression based on the audio and video features.
They released the dataset DAIC-WOZ, which contains video recordings of clinical interviews designed to diagnose psychological disorders.
The user participated in the conversation after completing a depression self-diagnosis questionnaire.
DAIC-WOZ has significantly advanced depression detection research, contributing to numerous breakthroughs in the field \citep{he2018automated, haque2018measuring, low2020automated}.
We utilize DAIC-WOZ as a benchmark due to the absence of text-based depression diagnosis conversation datasets.

\subsection{Dialogue Data Synthesis}
Several methodologies have been proposed for the generation and augmentation of dialogue data to address the constraints associated with the time-intensive and costly data construction process \citep{lewis2017deal, hou2018sequence, tang2019target}.
With the emergence of the Large Language Model (LLM), the field of data synthesis has transitioned into a novel paradigm \citep{ding2024data}.

\begin{figure*}[t!]
  \centering
  \includegraphics[width=\linewidth]{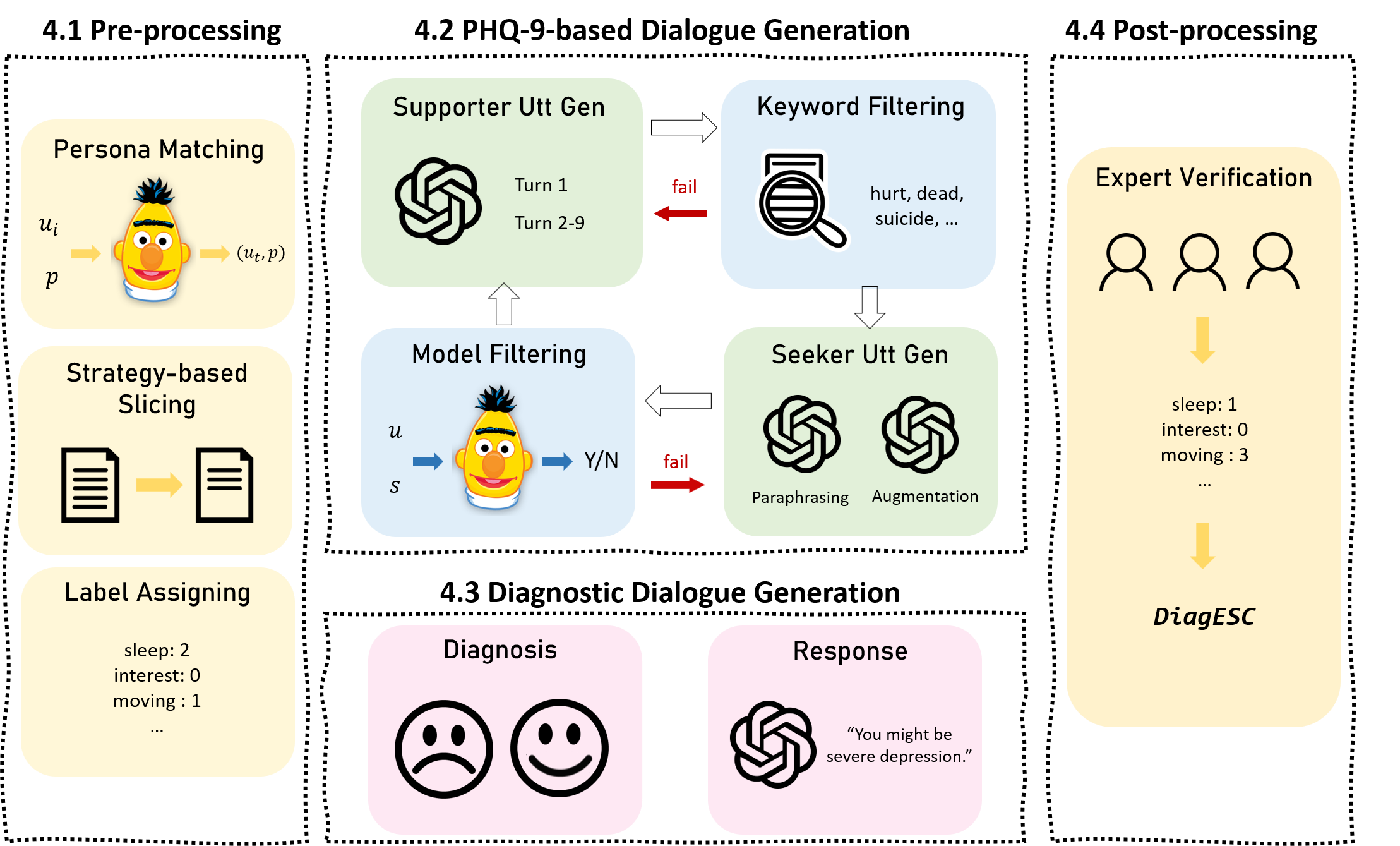}
  \caption{The overview of the DESC synthesis process.}
  \label{fig:process}
\end{figure*}

\citet{kim2023soda, bao-etal-2023-synthetic} introduced a novel synthetic dialogue dataset derived from external sources.
The data was refined using filtering techniques designed to ensure criteria such as commonsense knowledge, dialogue flow, and coherence.
A method for synthesizing Dialogue State Tracking (DST) labeled conversation data from dialogue schemas and templates has shown comparable performance to human-annotated datasets in few-shot DST \citep{kulkarni-etal-2024-synthdst}.
\citet{kim2024kocosa, li2024stylechat} generated the conversational dataset using task-specific prompting technology, and the test set is certificated through humans. 

Building on these advancements, we synthesize the DESC dataset by leveraging the fluent utterance generation capabilities of LLM, thereby contributing a novel resource for text-based depression detection research.

\section{Problem Formulation}
DiagESC consists of three sub-tasks—two modes response generation, persona generation, and diagnosis generation.
The dual modes of response generation encompass emotional support and diagnostic responses.
Persona generation is extracting characteristics based on the seeker's previous utterance, as suggested by PAL \citep{cheng2023pal}.
Utilizing the previous persona as input can increase the user experience by facilitating reflection on an individual's characteristics and serving as a form of memory when the dialogue history cannot include all utterances.
Diagnosis generation, introduced for DiagESC, involves generating symptom and corresponding score pairs.

Equation~\ref{eq:task} refers to DiagESC $\mathcal{F}$ at turn $t$ that generates response $r_t=m\oplus s_t$, persona $p_t$, and diagnosis pair $d_t=\{\mathrm{Symptom}, \mathrm{Score}\}$ given the persona sentences $P_t=\{p_1, p_2, \ldots, p_t\}$ and dialogue history $C_t = \{u_1, s_1, u_2, s_2, \ldots, u_t\}$.
$m\in \{\mathrm{emotional\,support}, \mathrm{diagnostic}\}$ denotes the mode for response.
$u_t$ and $s_t$ represent $t^{\text{th}}$ seeker (user) and supporter (system) utterance respectively.
\begin{equation}
    \mathcal{F}(P_t, C_t) = (r_t, p_t, d_t)
\label{eq:task}
\end{equation}x
The $\mathrm{emotional\,support}$ response uses existing ESC strategies, and the $\mathrm{diagnostic}$ response includes depression symptom questions and the notification diagnosis result.

\section{Methodology}
We synthesize the dataset for DiagESC named DESC through a four-step process, as illustrated in Figure~\ref{fig:process}.
Initially, the source data undergoes pre-processing to align with the task requirements (Section~\ref{sec:pre}).
PHQ-9-based Dialogue Generation is for generating conversations that ask and answer about symptoms of depression (Section~\ref{sec:phq}).
Each task-specific prompt is based on the Patient Health Questionnaire-9 (PHQ-9) symptom item.
The process includes filtering for reliability.
The severity of depression is then calculated based on the answers obtained by the seeker.
Section~\ref{sec:diag} is to inform the seeker of appropriate advice.
Finally, to enhance reliability, expert verification is conducted on the validation and test datasets (Section~\ref{sec:post}).

\begin{table*}[t]
\scalebox{0.85}{
\begin{tabular}{c|m{15cm}}
\hline
\textbf{Symptom Item} & \textbf{Description}                                                                                                                                                    \\ \hline 
Interest              & Little interest or pleasure in doing things                                                                                                                             \\ 
Depressed             & Feeling down, depressed, or hopeless                                                                                                                                    \\ 
Sleep                 & Trouble falling or staying asleep, or sleeping too much                                                                                                                 \\ 
Tired                 & Feeling tired or having little energy                                                                                                                                   \\ 
Appetite              & Poor appetite or overeating                                                                                                                                             \\  
Failure               & Feeling you are a failure or have let yourself or your family down                                                                                                      \\ 
Concentrating         & Trouble concentrating on things, such as reading the newspaper or watching television                                                                                   \\
Moving                & Moving or speaking so slowly that other people could have noticed. Or the opposite, being so fidgety or restless that you have been moving around a lot more than usual \\ 
Hurting               & Thoughts that you would be better off dead, or of hurting yourself       \\ \hline                                                       
\end{tabular}
}
\caption{The symptoms and descriptions of PHQ-9}
\label{tab:phqitem}
\end{table*}
\subsection{Pre-processing}
\label{sec:pre}
\noindent
\textbf{Persona Matching}
We utilize comprehensive annotations and high-quality supporting dialogue from the PESConv dataset of PAL \citep{cheng2023pal}, containing persona sentences extracted from previous dialogue history.
However, the persona sentences do not align exactly with the seeker's utterance for each turn.

We employ the BERT\footnote{https://huggingface.co/google-bert/bert-base-uncased} \citep{kenton2019bert} model to obtain the embeddings for all persona sentences $p$ and seeker utterances $u$.
Then, we compute cosine similarities between each persona sentence embedding and every utterance embedding.
Each persona sentence has a higher cosine similarity to the utterance from which it is derived than to other utterances.
We reassign all of the persona sentences using the following equation.
\begin{equation}
\label{eq:sim}
\hat{t_i}= \mathrm{argmax}_{t \in \{1, \cdots ,T\}} \frac{E(p_i) \cdot E(u_t)}{|E(p_i)| |E(u_t)|}
\end{equation}
where $\hat{t_i}$ represents the matched turn number for the $i$-th persona sentence $p_i$.
$u_t$ denotes the utterance at the $t$-th turn and the function $E(\cdot)$ refers to compute BERT embedding.
Equation~\ref{eq:sim} ensures to align each persona sentence with its derived utterance.

\noindent
\textbf{Strategy-based Slicing}
Determining the appropriate moment to begin diagnostic questions is challenging.
It is crucial to consider that abruptly interrupting the flow of conversation may negatively impact the user's emotional state.
Fortunately, ESC dataset has rich annotations, tagging each utterance with its corresponding strategy.
The most suitable time for presenting diagnostic questions has been empirically determined to use specific strategies, namely \textit{Restatement or Paraphrasing}, \textit{Reflection of Feeling}, \textit{Self-disclosure}, and \textit{Affirmation and Reassurance}.
Figure~\ref{fig:sample} is an example of using a diagnostic question (yellow) instead of a reflection response (gray).
This rule enables a smooth and contextually appropriate transition into diagnostic questioning.

Furthermore, we only utilize truncated data when at least two persona sentences have been gathered to ensure that diagnostic questions are only posed after comprehensively understanding the seeker's persona.
This criterion helps that sufficient contextual background is considered before diagnostic engagement.

\noindent
\textbf{Label Assigning}
To achieve an even distribution of the final severity level within the generated conversational data, the PHQ-9 labels are pre-assigned.
The next step uses predefined labels to generate utterances.

\subsection{PHQ-9-based Dialogue Generation}
\label{sec:phq}
We utilize the Patient Health Questionnaire-9 (PHQ-9), a widely used medical tool for self-assessment of depression \citep{kroenke2001phq}, as the basis for the depression diagnostic questions.
PHQ-9 aims to quantify the frequency of nine depressive symptoms listed in Table~\ref{tab:phqitem} on a scale ranging from 0 to 3, with the options \textit{Not at all}, \textit{Several days}, \textit{More than half the days}, and \textit{Nearly every day}.
The aggregated score of all items is used to diagnose depression and assess its severity, categorized as Minimal (0-4), Mild (5-9), Moderate (10-14), Moderately severe (15-19), and Severe (20-27).

\subsubsection{Supporter Utterance Generation}
We develop two types of prompts to generate supporter utterances for the first and subsequent turns.
In the initial turn, it is essential to formulate questions with caution to maintain a positive user experience.
For the subsequent turns, which involve further diagnostic questioning, it becomes essential to comprehend and empathize with the seeker's responses to the preceding questions.

The both prompts involve the three-step Chain-of-Thought technique \citep{wei2022chain, kim2024kocosa}.
In the first-turn supporter utterance generation prompt, the steps consist of \textit{Selection}, \textit{Planning}, and \textit{Response Generation}.
Table~\ref{tab:prompt-sys1} provides detailed instructions for these steps.
\textit{Selection} and \textit{Planning} focus on the seeker's persona and the dialogue history.

\begin{table}[t]
\begin{tabular}{p{0.97\linewidth}}
\hline
\textbf{Prompt Content}
\\ \hline
You are an emotional supporter. You have to ask about the frequency of depression symptoms without compromising the emotions of the user suspected of having depression. \\
\textbf{Depression Symptoms} You should ask how `often' a symptom has occurred over the past two weeks. Use one of the following symptoms. Be careful not to distort the medical meaning. \textit{(symptoms)}   \\
\textbf{Task Description} The task proceeds in three stages: Selection, Planning, and Response Generation. The first step, Selection, is to select which of the given persona sentences and dialog history to use in the response generation and what symptoms to ask about. Information that can improve the user experience must be extracted. The second step, Planning, is planning how to use the selected information. You must explain how you will use the information you have selected and why you have selected that information. The final step, Response Generation, uses the selected information to naturally ask the user about depression symptoms. Consistency with persona and history must not be broken. Questions must be asked carefully so that the user does not feel that the question is sudden. Be careful not to ask hard as if you were being interrogated. The generated response must be no more than 25 words.    
\\ 
\textbf{Example} \textit{(examples)}
\\ \hline
\end{tabular}
\caption{The prompt used to generate the initial supporter utterance of inquiring about PHQ-9 symptoms.}
\label{tab:prompt-sys1}
\end{table}

Because analyzing the previous answer about the symptom is more critical for subsequent turns, we replaced the \textit{Selection} with the \textit{Analysis}.
The steps help to analyze the seeker's response and generate a response accordingly.
Detailed instructions can be found in the Appendix~\ref{sec:appendix_prompts}.

\subsubsection{Seeker Utterance Generation}
\begin{figure}[t]
  \centering
  \includegraphics[width=0.9\columnwidth]{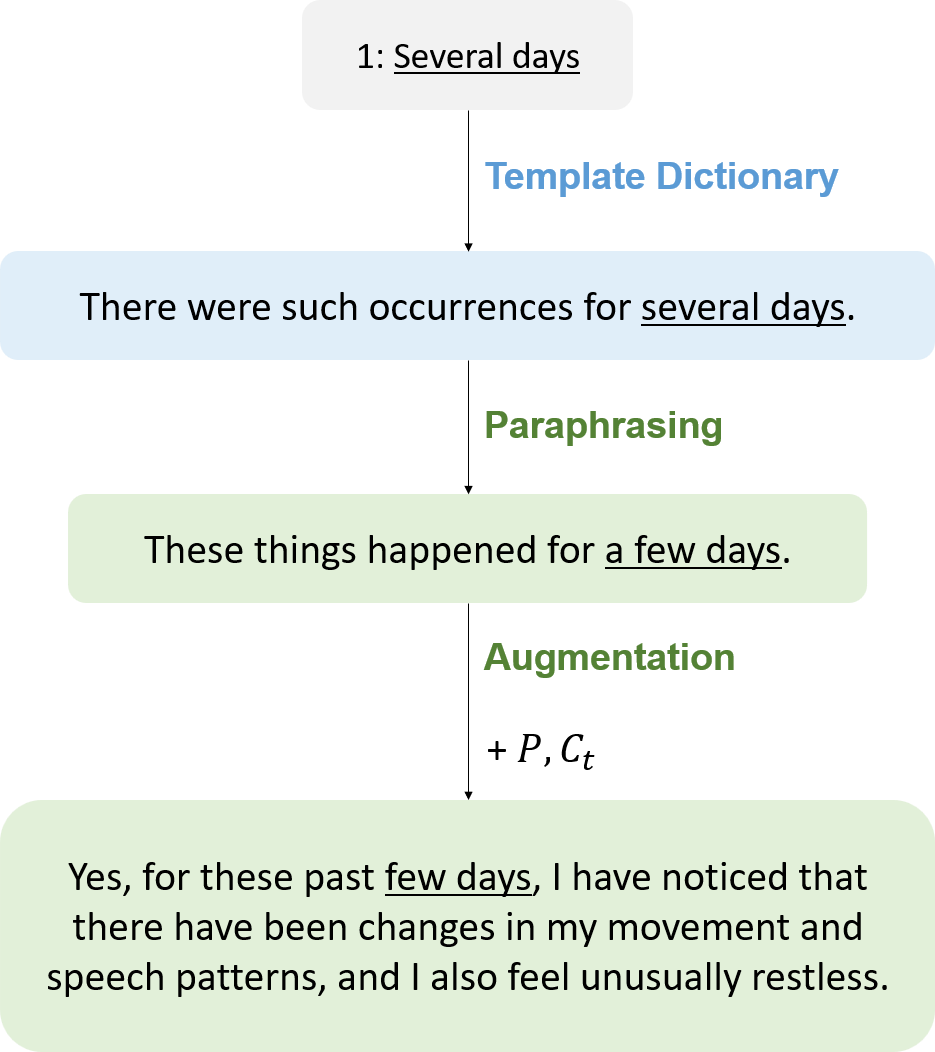}
  \caption{The overview of seeker utterance generation}
  \label{fig:userutt}
\end{figure}
\begin{table}[t]
\centering
\scalebox{0.75}{
\begin{tabular}{c|c|l}
\hline
Score              & Type  & \multicolumn{1}{c}{Content}                   \\ \hline
\multirow{2}{*}{0} & PHQ   & Not at all                                    \\
                   & Temp. & There is not at all much of that.             \\ \hline
\multirow{2}{*}{1} & PHQ   & Several days                                  \\
                   & Temp. & There were such occurrences for several days. \\ \hline
\multirow{2}{*}{2} & PHQ   & More that half the days                       \\
                   & Temp. & That happened for more than half the days.    \\ \hline
\multirow{2}{*}{3} & PHQ   & Nearly every day                              \\
                   & Temp. & It was like that almost every day.           \\ \hline
\end{tabular}
}
\caption{Templates by score used for the Paraphrasing stage in generating seeker utterances.}
\label{tab:template}
\end{table}
The construction of the seeker's utterance necessitates including one of the four designated responses from PHQ-9 (Not at all, Several days, More than half the days, Nearly every day).
We enhance and utilize the template-based utterance generation method \citep{kulkarni-etal-2024-synthdst}.
After rephrasing the templates, we augment the utterances to match the user's characteristics better, personalizing them to align more closely with specific traits and preferences, providing a more tailored and engaging conversational experience.

Figure \ref{fig:userutt} shows the process by which the template is transformed into the final utterance via the stages of Paraphrasing and Augmentation using LLM.
Initially, we establish a foundational template that directly correlates with a response option from the PHQ-9, as illustrated in Table \ref{tab:template}.

Paraphrasing model $\mathbf{M}_{\mathrm{Para}}$ aims to diversify responses while preserving symptom frequency information.
Augmentation model $\mathbf{M}_{\mathrm{Aug}}$ represents the process of generating the final utterance $u_t$ to align with the seeker's specific situation and persona, particularly considering the supporter's last utterance and the overall conversational flow.
As can be seen in Equation~\ref{eq:user}, $\mathbf{M}_{\mathrm{Para}}$ requires a template of the corresponding score $s$ as input and $\mathbf{M}_{\mathrm{Aug}}$ demands a paraphrased output, the persona sentences $P_t$, and the conversational history $C_t$.
\begin{equation}
    u_t = \mathbf{M}_{\mathrm{Aug}}(P_t, C_t, \mathbf{M}_{\mathrm{Para}}(\mathrm{template}_s))
\label{eq:user}
\end{equation}

\subsubsection{Filtering}
If the LLM misrepresents the medical interpretation of PHQ-9, it will adversely affect the quality of the diagnosis function.
We implement strict filtering algorithms to prevent the hallucination of LLM and ensure the reliability of diagnostic conversations.

\noindent
\textbf{Keyword Filtering}
To maintain the integrity of symptom representation in LLM-generated supporter utterances, we pre-define specific keywords associated with each symptom.
If each utterance includes no pre-defined keywords, we discard it and regenerate a new one.
This process continues until the generated utterance appropriately incorporates the necessary keywords, ensuring accurate and consistent symptom representation in the dialogue.

\noindent
\textbf{Model Filtering}
It is essential to preserve symptom frequency information in the seeker's utterances.
Therefore, we train the classification model $M_c$, using 256 manually verified utterances.
If the predicted label from model $M_c$ differs from the template label, or if the confidence is below a threshold $t$, the utterance is regenerated.

\subsection{Diagnostic Dialogue Generation}
\label{sec:diag}
The final goal of DiagESC's diagnostic ability is to estimate the seeker's mental health status and provide appropriate assistance.
The severity level of depression is determined by summing the scores of all nine items obtained from the Diagnosis task and then generating an appropriate response as shown in the final utterance in Figure~\ref{fig:sample}.
To achieve this goal, we design the prompt in Table~\ref{tab:last} to generate an utterance based on the seeker's persona and a diagnosed depression severity level.
To enhance the naturalness of the conversation, we incorporate a predefined turn expressing gratitude for the honest response before the diagnostic response.

\begin{table}[t]
\begin{tabular}{p{0.98\linewidth}}
\hline
\textbf{Prompt Content}
\\ \hline
You are emotional support. You have provided counseling to the user about the concerns and even completed questions about depression symptoms. Generate an utterance that concludes the counseling by referring to the depression diagnosis results and the user's persona. If the severity of depression is high, you should be recommended to see a hospital or counselor. Please generate the utterance friendly conversational style and generated utterance must be no more than 30 words. 
\\ 
\textbf{Example} \textit{(examples)}
\\ \hline
\end{tabular}
\caption{The prompt used to generate utterance for notifying diagnosis result.}
\label{tab:last}
\end{table}

\subsection{Post-processing}
\label{sec:post}
Despite applying strict task-specific filtering protocols, the potential for inaccuracies remains.
To ensure the reliability of the PHQ-9 labels, Expert Filtering is conducted on the validation and test sets of DESC.
Three psychologists, who are native English speakers or bilingual and have over four years of professional experience\footnote{We hired psychologists through https://www.upwork.com}, assessed scores for each symptom.
The seeker utterances are then re-labeled to the mode value of the three scores.

\section{Experiments}
\subsection{Diagnostic Ability Evaluation}
The DAIC-WOZ dataset, used as a baseline, comprises clinical dialogues in video and audio features with PHQ-8 \citep{kroenke2009phq} labels.
The PHQ-8 is a modified version of the PHQ-9, excluding the items related to suicide, and performs just as well as the PHQ-9 in diagnosing depression.
Although the modality is different from ours, due to the absence of conversation data explicitly labeled for depression, we use transcripts of DAIC-WOZ.
We randomly sample four dialogues from DAIC-WOZ for each severity level.
Employing the same methodology with expert labeling described in Section \ref{sec:post}, three psychological counselors evaluate scores for the PHQ-8 items.

The Quadratic Weighted Kappa (QWK) score is a metric that evaluates the agreement between two predictions, offering advantages by acknowledging both exact and partial alignment in assessments \citep{cohen1968weighted}.
The QWK score ranges from -1 to 1. 
A score closer to -1 indicates that the predictions are nearly opposite.
A score near 0 reflects randomness, implying no consistent agreement between the predictions.
If a score approaches 1, the predictions are almost identical and have a high level of agreement.
As it is suitable for medical fields where symptoms can be interpreted slightly differently depending on the individual doctor \citep{yoshida2015reproducibility, nirthika2020loss, 10045938}, the QWK score is widely used in disease diagnosis.
Therefore, we adopt the QWK score as the principal metric to evaluate diagnostic ability.

\begin{table*}[t]
\centering
\scalebox{0.85}{
\begin{tabular}{c|ccccccccc|c}
\hline
Dataset  & Interest & Depressed & Sleep & Tired & Appetite & Failure & Concentrating & Moving & Hurting & Avg         \\ \hline
DAIC-WOZ & 0.16     & 0.39      & 0.18  & 0.15  & 0.03     & 0.03    & -0.13         & 0.04   & -       & 0.11          \\
DESC  & 0.44     & 0.69      & 0.64  & 0.70  & 0.80     & 0.80    & 0.64          & 0.78   & 0.81    & \textbf{0.70} \\ \hline
\end{tabular}
}
\caption{Average QWK Scores of each dataset against expert annotations for each symptom in PHQ-8 and PHQ-9.}
\label{tab:qwk}
\end{table*}
\subsection{Conversational Quality Evaluation}
To evaluate the quality of the conversational data, we sample 10 dialogues for each severity level from the DESC validation and test sets.
We then requested the same evaluators with diagnostic ability evaluation to rate the following three items on a scale from 1 (Very Pool) to 5 (Excellent).
\begin{itemize}
    \item \textbf{Fluency} evaluates the grammatical correctness, naturalness, and smoothness of the dialog.
    \item \textbf{Consistency} assesses how well the dialogue maintains a consistent user persona throughout the interaction. This involves the user's interests and personality traits.
    \item \textbf{Coherence} measures how contextually appropriate the responses are, considering the previous dialogue turns and the overall context of the conversation.
\end{itemize}

\begin{figure}[t]
  \centering
  \includegraphics[width=0.8\columnwidth]{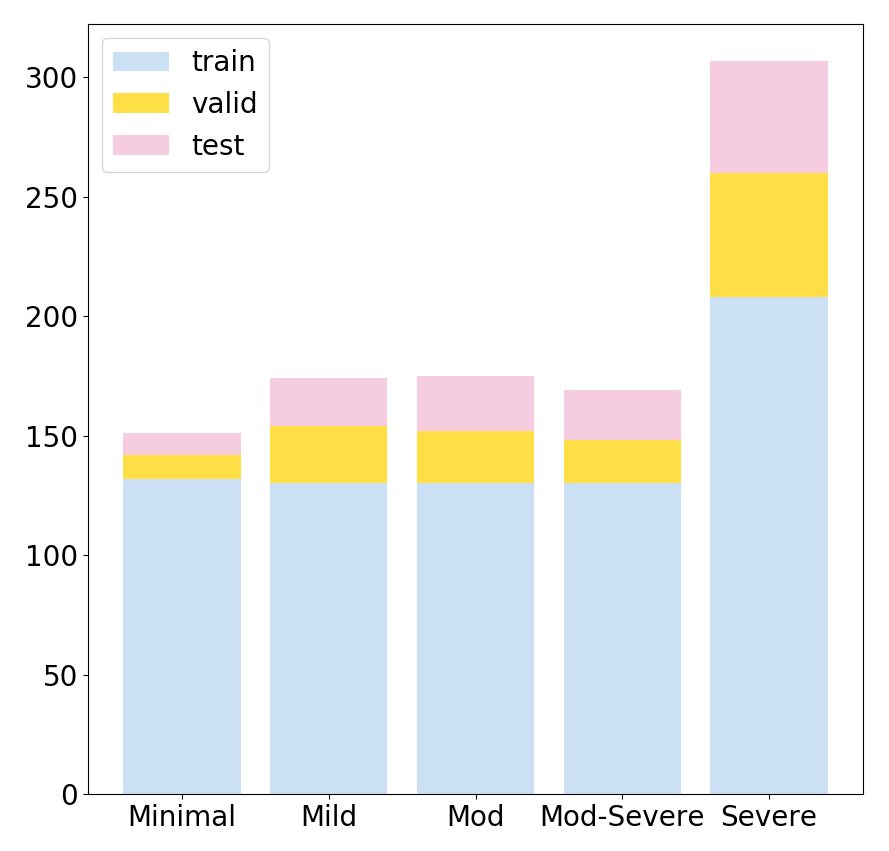}
  \caption{Distribution of depression severity labels in DESC. Minimal (0-4), Mild (5-9), Moderate (10-14), Moderately severe (15-19), and Severe (20-27).}
  \label{fig:severity_distribution}
\end{figure}

\subsection{Automatic Validation in Baseline}
We establish the baseline models by tuning a small LLM utilizing Low-Rank Adaptation (LoRA) \citep{hu2022lora} to evaluate the operation of DESC in each model.
We train the LoRA adapters on DESC for the three sub-tasks: response, persona, and diagnosis generation.
Additionally, performance in a multi-adapter setting is measured to evaluate the multi-tasking capabilities of the system.
In a single-task setting, the inputs for the next turn are the true labels of other tasks.
However, the inferred results from the previous turn are used as input for all tasks in a multi-task setting.
We systematically provide symptom item sequences to ensure consistency and effectiveness in the diagnostic process.

\subsection{Implementation Details}
We use GPT-4 as LLM to generate utterances of DESC.
For Model Filtering of seeker utterance, we adopt RoBERTa\footnote{FacebookAI/roberta-base} \citep{liu2019roberta} and train the classification model $M_c$ for 5 epochs.
The labels predicted by the fine-tuned model are utilized for filtering purposes, with the threshold $t=0.7$.
\begin{table}[t]
\scalebox{0.75}{
\begin{tabular}{c|c|cccc}
\hline
\multirow{2}{*}{Dataset} & \multirow{2}{*}{Level Acc} & \multicolumn{4}{c}{Depression}   \\
                         &                            & Acc  & Precision & Recall & F1   \\ \hline
DAIC-WOZ                 & 0.45                       & 0.70 & 0.50      & 1.00   & 0.67 \\
DESC                  & 0.71                       & 0.89 & 1.00      & 0.86   & 0.92 \\ \hline
\end{tabular}
}
\caption{Accuracy of predicting depression severity level and accuracy, precision, recall, and f1 score of estimating depression diagnosis.}
\label{tab:diagnosis}
\end{table}
Llama2\footnote{meta-llama/Llama-2-7b-chat-hf} \citep{touvron2023llama} is used as a baseline model, and the adapters are trained with the train set for 5 epochs on 4 NVIDIA A6000 GPUs, and the final model with the lowest validation loss was selected.
We employ AdamW with a learning rate of 5e-5 and a linear scheduler.

\section{Results and Analysis}
\subsection{Basic Statistics of DESC}
The DESC comprises 976 dialogues, including 730 train, 126 validation, and 120 test samples.
Each dialogue has an average of 42 turns, with the maximum number of turns per dialogue being 111 and the minimum being 24.
Figure~\ref{fig:severity_distribution} illustrates the distribution of dialogue samples across five levels of depression severity.
The Severe level has more samples than the other levels because it covers a wider range of scores.

\subsection{Diagnosis Ability}
According to the results presented in Table~\ref{tab:qwk}, DESC achieves a notably high average QWK score of 0.70 compared to baseline.
In contrast, DAIC-WOZ obtains low scores, with a certain item showing negative values.
The result indicates a substantial challenge in predicting the frequency of a seeker's depression symptoms solely from conversational history with an agent in the dataset. 
This comparison may be considered unfair because the DAIC-WOZ does not include questions about all the symptoms of the PHQ-8.

The most important result is the final depression diagnostic capability of each dataset, as presented in Table~\ref{tab:diagnosis}.
The depression severity is classified into five levels—minimal, mild, moderate, moderately severe, and severe—based on the cumulative scores of the assessed items.
Scores exceeding 10 points, classified as moderate or higher, are considered depression.

The Level Acc indicates the accuracy of predicting severity levels, with DESC showing 0.26 higher performance than DAIC-WOZ.
In depression diagnosis, the superior accuracy and F1 score of DESC, compared to the baseline, demonstrate its robustness and effectiveness.
The results suggest that our PHQ-9-based data generation process ensures reliable diagnostic capabilities.

\subsection{Conversation Quality}
\begin{figure}[t]
  \centering
  \includegraphics[width=0.9\columnwidth]{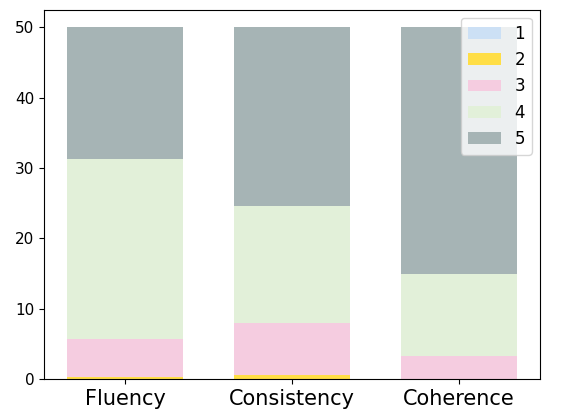}
  \caption{Distribution of evaluated scores for DESC’s fluency, consistency, and coherence.}
  \label{fig:quality}
\end{figure}

Figure~\ref{fig:quality} shows the distribution of obtained scores in the human evaluation performed to evaluate conversation quality.
The average scores are 4.25 for fluency, 4.33 for consistency, and 4.63 for coherence.
Most samples received scores of 3 or higher across all items, indicating that the DESC is consistent and comprises high-quality conversations without contextual awkwardness.
Notably, the high coherence score suggests that the diagnostic questions generated through strategy-based slicing and first-turn supporter prompt configuration help the seeker perceive them as natural and non-abrupt.

\subsection{Automatic Validation}
\begin{table}[t]
\centering
\begin{tabular}{c|cc|c|cl}
\hline
\multirow{2}{*}{} & \multicolumn{2}{c|}{Response} & Persona & \multicolumn{2}{c}{Diagnosis} \\
                  & Mode          & BLEU          & BLEU    & \multicolumn{2}{c}{Acc}       \\ \hline
Single            & 0.83          & 31.08         & 34.03   & \multicolumn{2}{c}{0.78}      \\ 
Multi             & 0.83          & 30.78         & 34.65   & \multicolumn{2}{c}{0.77}     
\\ \hline
\end{tabular}
\caption{The performance on baseline models for single-task and multi-task settings.}
\label{tab:model}
\end{table}
Table~\ref{tab:model} shows the baseline performance of DESC.
Mode indicates the prediction accuracy of the response mode, divided into emotional support and diagnosis.
It shows equal performance of 0.83 in both single-task and multi-task settings.
Generating response and persona sentences achieve high BLEU scores, all above 30.
Diagnosis accuracy measures the prediction of each symptom and its corresponding score.
Across all metrics, single-task and multi-task settings demonstrate similar performance.

\section{Conclusion}
This work proposes the DiagESC task for a comprehensive mental health care dialogue system that goes beyond the limitations of supportive dialogue systems that do not detect mental risk.
DiagESC contributes to emotional support and early detection of depression, an important part of mental health.
We have released the novel dataset DESC by synthesizing diagnostic conversations based on a depression self-diagnosis questionnaire with emotional support data.
Task-specific prompts and strict filtering protocols facilitate questions about depression symptoms while ensuring continued user engagement.
Evaluation by a psychological counseling expert proves that DESC has superior diagnostic performance and conversational quality.
We hope that DiagESC will contribute significantly to developing more effective and supportive dialogue systems in mental health care.
Moreover, the release of the DESC dataset provides a valuable resource for the research community, encouraging further advancements and innovations in this critical area.

\section*{Limitation}
The research aims to identify depression signs during conversations with the user, subsequently notifying them of potential risks. It is imperative to note that the diagnostic outcomes derived from the proposed dataset and model are intended solely for guidance. An accurate and definitive diagnosis should be ascertained through consultation with a medical professional.

\section*{Acknowledgements}
This work was supported by Smart HealthCare Program(www.kipot.or.kr) funded by the Korean National Police Agency(KNPA, Korea) [Project Name: Development of an Intelligent Big Data Integrated Platform for Police Officers’ Personalized Healthcare / Project Number: 220222M01]
This research was supported by the MSIT(Ministry of Science and ICT), Korea, under the ITRC(Information Technology Research Center) support program(IITP-2024-RS-2024-00437866) supervised by the IITP(Institute for Information \& Communications Technology Planning \& Evaluation)

\bibliography{custom}

\appendix
\section{Diaolgue Example}
\begin{figure}[t]
    \centering
    \includegraphics[width=\columnwidth]{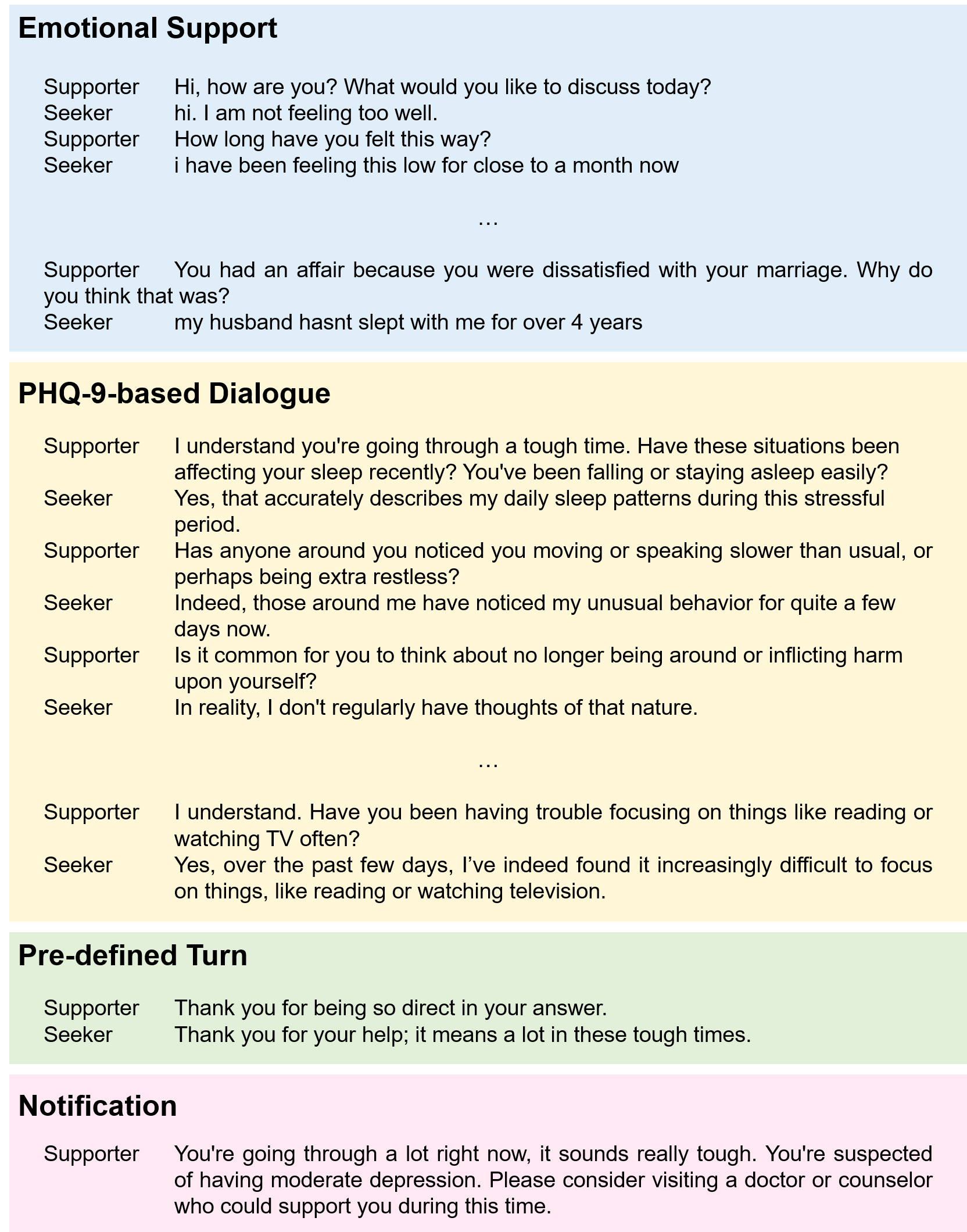}
    \caption{Part of an example conversation sample of DESC.}
  \label{fig:sample2}
\end{figure}
Figure~\ref{fig:sample2} illustrates the utterance configuration of the DESC data sample.
During emotional support conversation, the supporter initiates the PHQ-9-based dialogue to inquire about symptoms.
After identifying all symptoms, the supporter provides the user with appropriate advice, such as a recommendation to visit a hospital based on the diagnostic results.
To enhance the natural flow of the conversation, we insert the predefined turns between symptom inquiries and result notifications.
We select these turns from 23 supporter and 16 seeker utterance candidates.

\section{Detailed Prompt Instructions}
\label{sec:appendix_prompts}
\begin{table}[ht]
\begin{tabular}{p{0.98\linewidth}}
\hline
\textbf{Prompt Content}
\\ \hline
\textit{(same with turn 1)}
\\
\textbf{Depression Symptoms} You should ask how `often' a symptom has occurred over the past two weeks. Symptoms are given and the frequency of the symptoms should be naturally asked of the user. The meaning of a given symptom should never be changed.
\\
\textbf{Task Description} The task proceeds in three stages: Analysis, Planning, and Response Generation. The first step, Analysis, is to determine the user’s status through the user’s previous responses. The second step, Planning, is planning how to use the status information to support the user's emotions and ask about the frequency of a different given symptom. The final step, Response Generation, is to ask the user about the symptom according to plan. Question must be asked carefully so that the user does not feel that the question is sudden. Be ... \textit{(same with turn 1)}
\\ 
\textbf{Example} \textit{(examples)}
\\ \hline
\end{tabular}
\caption{The prompt used to generate the subsequent turn supporter utterance of inquiring about PHQ-9 symptoms.}
\label{tab:prompt-sys2}
\end{table}
\begin{table}[ht]
\begin{tabular}{p{0.98\linewidth}}
\hline
\textbf{Prompt Content}
\\ \hline
Rephrase the sentence while retaining the original meaning. The sentences are conversation with counseling diagnosis chatbot system and the user. In particular, do not change the frequency-related meaning of the user's words. Use synonyms or related words to express the sentences with the same meaning. Use conversational language and paraphrase the following sentences. Generate a crisp and to the point single sentence from the given sentences using conversational language.
\\ \hline
\end{tabular}
\caption{The Paraphrasing prompt used in seeker utterance generation}
\label{tab:user-para}
\end{table}

\begin{table}[ht]
\begin{tabular}{p{0.98\linewidth}}
\hline
\textbf{Prompt Content}
\\ \hline
Please augment the user utterance to fit the dialog history while maintaining its original meaning. The sentence is the user's utterance in a conversation between the counseling diagnosis chatbot system and the user. In particular, do not change the frequency-related meaning of user's words. Please augment and modify the given user utterance to match the system's last words and the flow of the conversation, especially user's situation and persona.
\\ \hline
\end{tabular}
\caption{The Augmentation prompt used in seeker utterance generation}
\label{tab:user-aug}
\end{table}

Table \ref{tab:prompt-sys2} is the prompt for supporter utterance generation.
In contrast to the utterance generation of turn 1, this stage focuses on analyzing the prior seeker response.
Table \ref{tab:user-para} and \ref{tab:user-aug} are the prompts for seeker utterance generation.

\section{Keyword Filtering}
\begin{table}[h]
    \centering
    \resizebox{\columnwidth}{!}{    
    \begin{tabular}{l|l}
    \hline
    Item          & Keywords                                           \\
    \hline
    Interest      & interest, pleasure, enjoy                         \\
    Depressed     & depressed, hopeless, down                           \\
    Sleep         & sleep                                             \\
    Tired         & tired, energy                                     \\
    Appetite      & appetite, eat                                     \\
    Failure       & fail, down                                        \\
    Concentrating & concentrate, concentrating, TV, television, read  \\
    Moving        & move, moving, slow, restless, figety              \\
    Hurting       & hurt, dead, suicide, self, harm                  
    \\ \hline
    \end{tabular}
    }
    \caption{The keyword list of each symptom item.}
    \label{tab:keyword}
\end{table}
The keyword filtering process ensures that the PHQ-9 maintains its medical meaning.
Table \ref{tab:keyword} shows detailed keywords for each symptom item.
The generated utterance must contain at least one of the keywords.

\section{Distribution of DESC}
\begin{figure}[h]
    \centering
    \includegraphics[width=\columnwidth]{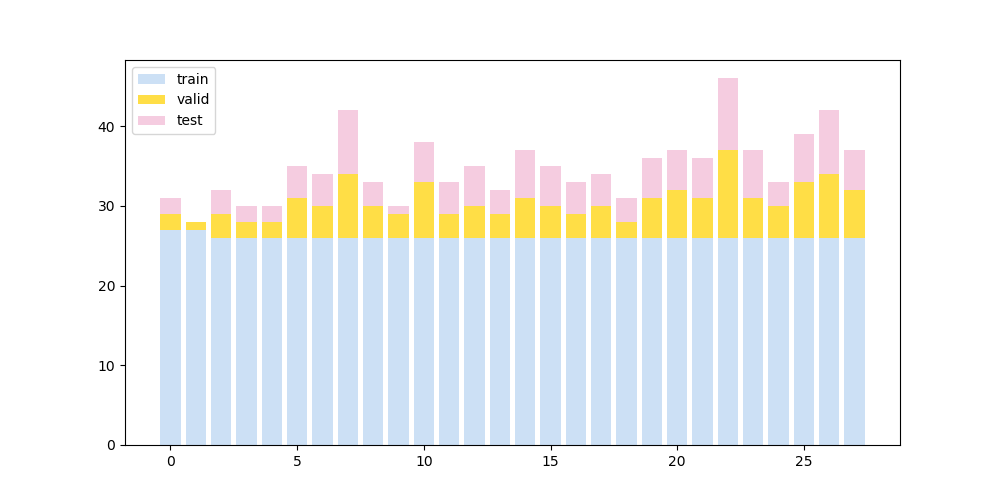}
    \caption{Distribution of aggregated score in DESC.}
  \label{fig:score_dist}
\end{figure}

Each dialogue sample has score labels for the PHQ-9 items.
The sum of all nine scores determines the severity of depression.
Figure \ref{fig:score_dist} is the distribution of aggregated scores.

\section{Human Evaluation}
We conduct two types of human evaluations.
In the diagnostic ability evaluation, three evaluators, all psychologists, read the conversations and scored each item of the PHQ-9.
We use the mode value as the final label to reduce individual subjectivity.
If the evaluators' scores differ, we use the mean score as the final label.

The conversation quality evaluation assesses performance based on fluency, consistency, and coherence scores.
The evaluators read the dialogues and assign a score between 1 and 5 for each criterion, following the descriptions provided for each item.

Fluency evaluates the grammatical correctness, naturalness, and smoothness of the dialogue.
\begin{itemize}
    \item \textbf{Very poor} numerous errors, hard to understand.
    \item \textbf{Poor} lacks smoothness but can be followed with some effort.
    \item \textbf{Normal} a natural rhythm to the conversation despite occasional awkwardness.
    \item \textbf{Good} natural, easy to follow.
    \item \textbf{Excellent} natural, grammatically sound and logically structured.
\end{itemize}
Consistency assesses how well the dialogue maintains a consistent user persona throughout the interaction. This involves the user's interests, and personality traits.
\begin{itemize}
    \item \textbf{Very poor} frequent contradictory utterances; feels like by a completely different person.
    \item \textbf{Poor} regular contradictory utterances; a general sense of the original character remains perceivable but persona seem inconsistent.
    \item \textbf{Normal} some contradictory utterances; occasional contradictory utterances that mildly affect the coherence of the user persona but do not substantially alter the overall character impression.
    \item \textbf{Good} few errors; it's pretty much the same person speaking.
    \item \textbf{Excellent} no or negligible errors; user personas are fully maintained.
\end{itemize}
Coherency measures how contextually appropriate the responses are, considering the previous dialogue turns and the overall context of the conversation.
\begin{itemize}
    \item \textbf{Very poor} conversations frequently veer off-topic without a clear reason.
    \item \textbf{Poor} related to main topic but may include irrelevant details.
    \item \textbf{Normal} related topic with occasional lapses in focus or clarity.
    \item \textbf{Good} related topic and minor deviations are quickly corrected.
    \item \textbf{Excellent} every response directly contributes to a coherent, logical, and engaging.
\end{itemize}

\end{document}